# Fuzzy Approximate Reasoning Method based on Least Common Multiple and Its Property Analysis

I.M. Son, S.I. Kwak*, M.O. Choe

**Abstract.** This paper shows a novel fuzzy approximate reasoning method based on the least common multiple (LCM). Its fundamental idea is to obtain a new fuzzy reasoning result by the extended distance measure based on LCM between the antecedent fuzzy set and the consequent one in discrete SISO fuzzy system. The proposed method is called LCM one. And then this paper analyzes its some properties, i.e., the reductive property, information loss occurred in reasoning process, and the convergence of fuzzy control. Theoretical and experimental research results highlight that proposed method meaningfully improve the reductive property and information loss and controllability than the previous fuzzy reasoning methods.

**Keywords:** SISO fuzzy system: Fuzzy Approximate Reasoning: Least Common Multiple: Fuzzy Modus Ponens: Fuzzy Modus Tollens: Reductive Property: Fuzzy Controllability: Information Loss

## 1. Introduction

The fuzzy approximate reasoning is one of an important research branch in the uncertainty inference of Artificial Intelligence (AI) and Computational Intelligence (AI). This fuzzy approximate reasoning is consisted of two part, i.e., fuzzy modus ponens (FMP) and fuzzy modus tollens(FMP). In 1975 [38], the pioneer Lotfi A. Zadeh of fuzzy set theory, firstly in the world, proposed compositional rule of inference (CRI) for solving of FMP and FMT. There are a lot of models of fuzzy approximate reasoning in fuzzy system theory and various applications. And the reductive property is one of the essential and important properties in the approximate reasoning theory and it's applications. [3,16,23,34,36,37] However many researched fuzzy approximate reasoning methods have some shortcomings [1,2,4,9,24]. For example, as presented in [22,23], the underlying semantic of CRI [38] is linguistically unclear, and its fuzzy reasoning result does not completely satisfy the reductive property. In order to overcome the shortcoming of CRI, the triple implication principle (TIP) [32] was proposed. However as pointed out in [10], shortcoming of TIP method is that it cannot be applied in practical problem, for example, fuzzy control [10,11,15,16,25,31,32,36,37,39]. As presented in [22], the several fuzzy reasoning methods based on the fuzzy relation have a contradiction that they can be applied to the practical problem, for example fuzzy control, but do not satisfy the reductive property, vice versa.

Somewhat, in order to overcome the shortcoming of CRI and TIP, a lot of fuzzy reasoning methods based on similarity measure(SM) are proposed [24,25,27–29,34,35]. Their basic idea is to consider the similarity measure between the consequent $B(y)$ (resp. $A(x)$) and the fuzzy reasoning conclusion $B^*(y)$ (resp. $A^*(x)$) if the antecedent $A(x)$ (resp. $B(y)$) is similar to the given premise $A^*(x)$ (resp. $B^*(y)$) for FMP(resp. FMT). This idea is right. But, as mentioned in [30], the fuzzy reasoning methods based on SM depend strongly on the similarity measure and the modification function, and do not completely satisfy the reductive property. As presented in [12,13], due to many fuzzy reasoning methods based on SM do use nonlinear operators, the fuzzy sets of reasoning result are non–normal and non–convex ones. Therefore in fuzzy reasoning processing, linear operators must possibly be used. According to [2,8,18–21,30,33], a lot of fuzzy reasoning methods mathematically seem that they are all accompanied with a common shortcoming, that is, information loss. One of the reasons that do not satisfy the reductive property is to occur to losses of information occurred in reasoning processes. Therefore, information loss must possibly be reduced in fuzzy reasoning processing. And when fuzzy reasoning methodology is studied, the axiom that FMT is opposite to FMP must be considered.

In [40], in order to overcome shortcoming of the CRI and TIP, authors proposed a new Quintuple Implication Principle (QIP) for fuzzy reasoning, which draws the approximate reasoning conclusion $B^*(y)$ of the consequent $B(y)$ (resp. conclusion $A^*(x)$ of consequent $A(x)$) as the formula which is best supported by $A(x) \to B(y), A^*(x) \to A(x)$, and $A^*(x)$ (resp. $A(x) \to B(y), B(y) \to B^*(y)$, and $B^*(y)$), for FMP(resp. FMT). The proposed QIP was illustratively compared with the CRI and TIP solutions, which is much closer to the proposition that for example ''x is small'' and, therefore, in accordance with human thinking than CRI and TIP. In order to overcome the various shortcomings presented in [2,7,8,10,17–23,27,28,30,36, and 40], a fuzzy reasoning method based on new idea has proposed in [12,13], which has not some information losses due to use liner operators and standardizationization operation. That is, in [12,13], authors proposed a new fuzzy reasoning method based on Turksen and Zhong's Euclidian Distance Measure (DM) [5,27,28], so called DMM, which consists of both FMP–DM, and FMT–DM. The idea of the paper [12,13] is based on the paper [14].

In this paper we try to suggest a kind of distance measure based fuzzy approximate reasoning method in the single input single output (SISO) fuzzy system with discrete fuzzy set vectors. For this, we apply extended distance measure by using an least common multiple, will call it LCM method. Concretely, LCM method is consisted of two part, i.e., FMP–LCM, and FMT–LCM. In this paper, we compare the reductive properties for 5 fuzzy reasoning methods with respect to FMP and FMT, which are CRI, TIP, QIP, AARS, and our proposed LCM method. And we analysis theoretically on the information loss occurred in reasoning process, and the convergence for fuzzy control of the fuzzy reasoning method. The theoretical and experimental results highlight that the proposed approximate reasoning method is comparatively clear and effective, and in accordance with human thinking than other fuzzy reasoning methods.

The rest parts of this paper are organized as follows. In section 2, we discuss related works of distance measure and the method of FMP and FMT. And we show about the evaluation based on the criterion function of fuzzy reasoning methods. In section 3, a novel fuzzy approximate reasoning method for FMP and FM are proposed, i.e., FMP–LCM, and FM–LCM, respectively, and then those reductive properties are proved, and examples are showed for the proposed method. And in section 4, we discuss on the control property based on several fuzzy reasoning methods. In section 5, the reductive properties of Sade's CRI based method [38], Wang's TIP based method [32], Zhou et al.'s the QUIP based method [40], Turkmen et al.'s AIRS method [27,28], and our proposed LCM method for FMP and FM, are compared illustratively with respect to the reductive property. And then we consider on experiment results of the fuzzy controllability. In section 6, we describe the conclusion of this paper.

## 2. Related Works

Generally well known fuzzy reasoning methods are FMP and FM in the fuzzy system with 1 input 1 output 1 rule. General form of the fuzzy modus ponens presented in [7] is expressed as follows.

$$\text{Rule; if } x \text{ is } A \text{ then } y \text{ is } B, \quad \text{Premise}: x \text{ is } A^*, \quad \text{Conclusion}: y \text{ is } B^* \tag{1}$$

General form of FMT presented in [7] is also described as follows.

$$\text{Rule; if } x \text{ is } A \text{ then } y \text{ is } B, \quad \text{Premise}: y \text{ is } B^*, \quad \text{Conclusion}: x \text{ is } A^* \tag{2}$$

, where $A^* \in F(X)$, $A \in F(X)$ are fuzzy sets defined in the universe of discourse $X$, $B^* \in F(Y)$, $B \in F(Y)$ are fuzzy sets defined in the universe of discourse $Y$.

According to [7], the formula (2) can be written as follows, because FMT is opposite to FMP.

$$\text{Rule; if } y \text{ is } \overline{B} \text{ then } x \text{ is } \overline{A}, \quad \text{Premise}: y \text{ is } \overline{B}, \quad \text{Conclusion}: x \text{ is } \overline{A} \tag{3}$$

, where $\overline{A} = 1 - A$, $\overline{B} = 1 - B$. The most general forms of the CRI solutions of FMP and FMT are as follows.

$$\text{(FMP–CRI);} \quad B^*(y) = \bigvee_{x \in U} (A^*(x) \otimes (A(x) \to B(y))) \tag{4}$$

$$\text{(FMT–CRI);} \quad A^*(x) = \bigvee_{y \in V} (B^*(y)) \otimes (A(x) \to B(y))) \tag{5}$$

Where $\otimes$ is a left continuous t–norm and $\to$ its residual. TIP solution of FMP and FMT are expressed as follows.

$$\text{(FMP–TIP);} \quad B^*(y) = \bigvee_{x \in U} (A^*(x) \otimes (A(x) \to B(y))) \tag{6}$$

$$\text{(FMP–TIP);} \quad A^*(x) = \bigwedge_{y \in V} ((A(x) \to B(y)) \to B^*(y))) \tag{7}$$

The QIP solution of FMP and FMT presented in [40] is described as follows.

$$\text{(FMP–QIP);} \quad B^*(y) = \bigvee_{x \in U} (A^*(x) \otimes (A^*(x) \to A(x)) \otimes (A(x) \to B(y))) \tag{8}$$

$$\text{(FMP–QIP);} \quad A^*(x) = \bigvee_{y \in V} (A(x) \otimes (A(x) \to B(y)) \otimes (B(y) \to B^*(y))) \tag{9}$$

$$\text{(FMP-more or less form);} \quad B^*(y) = \min\{1, \ B(y)/SM(A^*(x), A(x))\} \tag{10}$$

$$\text{(FMT-more or less form);} \quad A^*(x) = \min\{1, \ A(x)/SM(B^*(y), B(y))\} \tag{11}$$

$$\text{(FMP-reduction form);} \quad B^*(y) = B(y) \times SM(A^*(x), A(x)) \tag{12}$$

$$\text{(FMT-reduction form)} \quad A^*(x) = A(x) \times SM(B^*(y), B(y)) \tag{13}$$

According to [28], one of distance measures (DM) for FMP, for FMT are as follows, respectively.

$$DM(A^*, A) = \left[\sum_{i=1}^{n}[\mu_{A^*}(x_i) - \mu_A(x_i)]^2 / n\right]^{1/2}, \text{ for FMP} \tag{14}$$

$$DM(B^*, B) = \left[\sum_{i=1}^{n}[\mu_{B^*}(y_i) - \mu_B(y_i)]^2 / n\right]^{1/2}, \text{ for FMT} \tag{15}$$

The $SM(A^*, A)$ based on $DM(A^*, A)$ is then defined as follows.

$$SM(A^*, A) = (1 + DM(A^*, A))^{-1} \text{ for FMP}, \quad SM(B^*, B) = (1 + DM(B^*, B))^{-1} \text{ for FMT} \tag{16}$$

Several existing similarity measure $SM(A^*, A)$ using Euclidian distance are as follows. [24,25,27–29,34,35]

$$SM(A, B) = 1 - \left(\sum_{u \in U} |\mu_A(u) - \mu_B(v)|\right) / \left(\sum_{u \in U} |\mu_A(u) + \mu_B(v)|\right) \tag{17}$$

$$SM(A, B) = \frac{1}{n}\sum_{i=1}^{n}[1 - \mu_A(u) - \mu_B(v)] \tag{18}$$

$$SM(A, B) = 1 - \max_{u \in U}(|\mu_A(u) - \mu_B(v)|) \tag{19}$$

$$SM(A, B) = \left[\sum_{u \in U}(\mu_A(u) \times \mu_B(v))\right] / \left[\max\left(\sum_{u \in U} \mu_A^2(u), \mu_B^2(v)\right)\right] \tag{20}$$

$$SM(A, B) = \sum_{u \in U} \min(\mu_A(u), \mu_B(v)) / \sum_{u \in U} \max(\mu_A(u), \mu_B(v)) \tag{21}$$

$$SM(A, B) = \frac{1}{n}\sum_{i=1}^{n}[\min(\mu_A(u), \mu_B(v)) / \max(\mu_A(u), \mu_B(v))] \tag{22}$$

$$SM(A, B) = \max\{\min(\mu_A(u), \mu_B(v))\} \tag{23}$$

$$SM(A, B) = \frac{1}{n}\sum_{i=1}^{n}\left\{\frac{1}{2}[(\mu_A(x_i) \leftrightarrow \mu_B(x_i)) + (\overline{\mu_A(x_i)} \leftrightarrow \overline{\mu_B(x_i)})]\right\},$$
$$\mu_A(\bullet) \leftrightarrow \mu_B(\bullet) = \min[(\mu_A(\bullet) \to \mu_B(\bullet)), (\mu_B(\bullet) \to \mu_A(\bullet))] \tag{24}$$

$$SM(A, B) = \sum_{i=1}^{n}[\mu_A(x_i) \times \mu_B(x_i) + v_A(x_i) \times v_B(x_i)] / \left\{\left(\sum_{i=1}^{n}[\mu_A^2(x_i) \times v_A^2(x_i)]\right) \times \left(\sum_{i=1}^{n}[\mu_B^2(x_i) \times v_B^2(x_i)]\right)\right\}^{\frac{1}{2}} \tag{25}$$

In the formula (25) $v_A(\bullet) = 1 - \mu_A(\bullet)$, $v_B(\bullet) = 1 - \mu_B(\bullet)$. For the evaluation for quality of the fuzzy reasoning result, the reductive property criteria for FMP and FMT based on [7,12,13,23] are shown in Table 1.

**Table 1.** Reductive property criteria for FMP and FMT based on [7,12,13,23]

| FMP | | *if x is A then y is B* |
|---|---|---|
| | Premise, $A^*$ | Reductive property criterion function $RPCF_{FMP}$ of $y$ $is$ $B^*$, (%) |
| Case 1 | $A^* = A$ | $RPCF_{FMP} = (1 - \sum_{k=1}^{\Theta}|b_{kl}^* - b_k|/\Theta) \times 100$ |
| Case 2 | $A^* = A^2$ | $RPCF_{FMP} = (1 - \sum_{k=1}^{\Theta}|b_{kl}^* - b_k^2|/\Theta) \times 100$ or $(1 - \sum_{k=1}^{\Theta}|b_{kl}^* - b_k|/\Theta) \times 100$ |
| Case 3 | $A^* = A^{1/2}$ | $RPCF_{FMP} = (1 - \sum_{k=1}^{\Theta}|b_{kl}^* - b_k^{\frac{1}{2}}|/\Theta) \times 100$ or $(1 - \sum_{k=1}^{\Theta}|b_{kl}^* - b_k|/\Theta) \times 100$ |
| Case 4 | $A^* = 1 - A$ | $RPCF_{FMP} = (1 - \sum_{k=1}^{\Theta}|b_{kl}^*(1 - b_k)|/\Theta) \times 100$ |
| Case 5 | $A^* = s.t.\ A$ | $RPCF_{FMP} = (1 - \sum_{k=1}^{\Theta}|b_{kl}^* - s.t.\ b_k|/\Theta) \times 100$ |
| FMT | | *if y is $\overline{B}$ then x is $\overline{A}$* |
| | Premise, $B^*$ | Reductive property criterion function $RPCF_{FMT}$ of $x$ $is$ $A^*$, (%) |
| Case 6 | $B^* = 1 - B$ | $RPCF_{FMT} = (1 - \sum_{k=1}^{\Theta}|a_{kl}^* - (1 - a_k)|/\Theta) \times 100$ |
| Case 7 | $B^* = 1 - B^2$ | $RPCF_{FMT} = (1 - \sum_{k=1}^{\Theta}|a_{kl}^* - (1 - a_k)^2|/\Theta) \times 100$ or $(1 - \sum_{k=1}^{\Theta}|a_{kl}^* - (1 - a_k)|/\Theta) \times 100$ |
| Case 8 | $B^* = 1 - B^{1/2}$ | $RPCF_{FMT} = (1 - \sum_{k=1}^{\Theta}|a_{kl}^* - (1 - a_k)^{1/2}|/\Theta) \times 100$ or $(1 - \sum_{k=1}^{\Theta}|a_{kl}^* - (1 - a_k)|/\Theta) \times 100$ |
| Case 9 | $B^* = B$ | $RPCF_{FMT} = (1 - \sum_{k=1}^{\Theta}|a_{kl}^* - a_k|/\Theta) \times 100$ |
| Case 10 | $B^* = s.t.\ B$ | $RPCF_{FMT} = (1 - \sum_{k=1}^{\Theta}|a_{kl}^* - s.t.\ a_k|/\Theta) \times 100$ |

In the Table 1, the indexes are $\Theta = lcm(u,v)$, least common multiple of $u$ and $v$, $u$ is number of fuzzy vector $a_{kl}^*$, $a_k$ and $v$ is number of fuzzy vector $b_{kl}^*$, $b_k$. And Class 1 contains Case 1, 2, 3, 4, for FMP, and Case 6, 7, 8, 9, for FMT, and Class 2 contains Case 1, 2, 3, 5, for FMP, and Case 6, 7, 8, 10, respectively. And s.t. is an abbreviated word of 《slightly tilted of》 mentioned in [7]. And $RPCF_{FMP}$, $RPCF_{FMT}$ are presented in [12, 13].

## 3. Proposed Fuzzy Approximate Reasoning Method

In this section, we propose a kind of distance measure based fuzzy approximate reasoning method in SISO fuzzy system with discrete fuzzy set vectors of different dimensions. This method is called an extended distance measure(LCM) based fuzzy approximate reasoning method, for short, LCM method. Our proposed LCM method is consisted of two part, i.e., FMP–LCM, and FMT–LCM. The fuzzy reasoning method that the dimension of the antecedent discrete fuzzy set is equal to one of the consequent discrete fuzzy set has already solved in the paper [12,13]. In this section, the discrete fuzzy set vectors of different dimensions mean that the dimension of the antecedent discrete fuzzy set differs from one of the consequent discrete fuzzy set in the SISO fuzzy system.

### 3.1. Fuzzy Approximate Reasoning Method For FMP–LCM

In this subsection, for the SISO fuzzy system with discrete fuzzy set vectors of different dimensions, we propose a novel fuzzy modus ponens based on least common multiple method, for short, FMP–LCM method, which differs from dimensions between the antecedent and consequent, i.e., mean the case of $u \neq v$ when element number of the antecedent is $u$ and element number of the consequent is $v$. A novel approximate reasoning method of FMP for SISO fuzzy system with discrete fuzzy set vectors of different dimensions is described as following Stages, which is so called FMP–LCM. A novel FMP-LCM is as follows.

**Stage 1**; Compute the extended fuzzy row vectors.

Let $u, v$ be all real integer index, then the extended fuzzy sets of the antecedent $A$, the given premise $A^*$ and consequent $B$, i.e., the least common multiple fuzzy vectors $\tilde{A}$, $\tilde{A}^* = \{\tilde{A}_l^*\}, \tilde{A}_l^* = [\tilde{a}_{1l}^*, \cdots, \tilde{a}_{il}^*, \cdots, \tilde{a}_{ul}^*] \, l = 1, 2, \cdots$, and $\tilde{B}$ with the index $\Theta = lcm(u,v)$ are calculated as following, respectively;

$$\tilde{A} = \left[\frac{a_1}{x_1}, \frac{a_2}{x_2}, \cdots, \frac{a_r}{x_r}, \cdots, \frac{a_{\Theta-1}}{x_{\Theta-1}}, \frac{a_\Theta}{x_\Theta}\right] \tag{26}$$

$$\tilde{A}_l^* = \left[\frac{a_{1l}^*}{x_1}, \frac{a_{2l}^*}{x_2}, \cdots, \frac{a_{rl}^*}{x_r}, \cdots, \frac{a_{(\Theta-1),l}^*}{x_{\Theta-1}}, \frac{a_{\Theta,l}^*}{x_\Theta}\right] \tag{27}$$

$$\tilde{B} = \left[\frac{b_1}{y_1}, \frac{b_2}{y_2}, \cdots, \frac{b_r}{y_r}, \cdots, \frac{b_{\Theta-1}}{y_{\Theta-1}}, \frac{b_\Theta}{y_\Theta}\right] \tag{28}$$

**Stage 2**; Compute the distance measure $LCM(\tilde{A}_l^*, \tilde{A})$ by least common multiple. Where index $l = 1, 2, \cdots$ means the number of the given premise fuzzy set, and $\Theta = lcm(u,v)$ is the least common multiple of $u$ and $v$. The distance measure $LCM(\tilde{A}_l^*, \tilde{A})_\Theta$, ($\Theta = lcm(u,v)$) is calculated as follows.

$$LCM(\tilde{A}_l^*, \tilde{A})_\Theta = \left[\frac{1}{\Theta}\sum_{r=1}^{\Theta}\left[a_{rl}^* - a_r\right]^2\right]^{1/2}, \text{ for FMP} \tag{29}$$

**Stage 3**; Compute the sign vectors $\tilde{P}_l$ by the difference $dif_{kl}$ of the given premise and the antecedent.

$$dif_{kl} = a_{kl}^* - a_k, (k = \overline{1, lcm(u,v)}, l = 1, 2, \cdots) \tag{30}$$

$$\tilde{P}_l = \left[\tilde{P}_{1l}, \tilde{P}_{2l}, \cdots \tilde{P}_{kl}, \cdots\right], \quad k = \overline{1, LCM(u,v)}, \quad l = 1, 2, \cdots \tag{31}$$

$$P(+1,0,-1)\text{form}; \tilde{P}_{kl} = sign(dif_{kl}) = \begin{cases} +1, & dif_{kl} > 0 \\ 0, & dif_{kl} = 0 \\ -1, & dif_{kl} < 0 \end{cases}, \text{for FMP–LCM} \tag{32}$$

$$P(+1,-1)\text{form}; \tilde{P}_{kl} = sign(dif_{kl}) = \begin{cases} +1, & dif_{kl} \geq 0 \\ -1, & dif_{kl} < 0 \end{cases}, \text{for FMP–LCM} \tag{33}$$

**Stage 4**; Compute the vectorialized distance measure $\tilde{C}_l$ since the distance measure $LCM(\tilde{A}_l^*, \tilde{A})_\Theta$ with an index $\Theta = lcm(u,v)$ is a scalar.

$$\tilde{C}_l = LCM(\tilde{A}_l^*, \tilde{A})_\Theta \times \tilde{P}_l, \ \Theta = lcm(u,v) \tag{34}$$

**Stage 5**; Obtain the quasi–quasi–approximate reasoning results $\tilde{B}_l^{**}, l=1,2,\cdots$ for FMP–LCM.

$$\tilde{B}_l^{**} = \begin{cases} \tilde{B}_l + \tilde{C}_l, & if \ \text{Case 1, 2, and 3} \\ 1 - \tilde{B}_l + \tilde{C}_l, & if \ \text{Case 4} \\ s.t. \ \tilde{B}_l + \tilde{C}_l, & if \ \text{Case 5} \end{cases} \tag{35}$$

**Stage 6**; Select the quasi–approximate reasoning results $B_l^{**}, l=1,2,\cdots$ from the quasi–quasi–approximate reasoning results $\tilde{B}_l^{**}$ for indexes $k=\overline{1, LCM(u,v)}$.

$$\tilde{B}_l^{**} \to B_l^{**}, \ i.e., \ \left[\tilde{b}_{lk}^{**}\right]_{\Theta \times 1} \to \left[b_{lq}^{**}\right]_{v \times 1} \tag{36}$$

Since the index $\Theta = lcm(u,v) = u \cdot m_1 = v \cdot m_2$, so $b_{lq}^{**} = \tilde{b}_{l,(q \cdot m_2)}^{**}$

**Stage 7**; Solve the individual approximate reasoning result $B_l^*$ from the quasi–approximate reasoning results $B_l^{**}$.

$$B_l^* = (B_l^{**} - \eta_l)/(\xi_l - \eta_l), l=1,2,\cdots \tag{37}$$

Where, the maximum $\xi_l$ and minimum $\eta_l$ of $B_l^{**}$ for FMP is calculated as follows.

$$\xi_l = \max B_l^{**}, \ \eta_l = \min B_l^{**}, l=1,2,\cdots. \tag{38}$$

**Stage 8**; For the SISO fuzzy system with discrete fuzzy set vectors of different dimensions, the final approximate reasoning result $B^*$ according to the given premises for FMP–LCM is obtained as follows.

$$B^* = \{B_l^*\}, \ l=1,2,\cdots \tag{39}$$

**Example 3.1** About the proposed FMP–LCM, let us consider the fuzzy system with $4 \times 1$ dimension antecedent fuzzy row vector $A(x)$=[1, 0.8, 0.4, 0] and $6 \times 1$ dimension consequent fuzzy row vector $B(y)$=[0, 0.2, 0.4, 0.7, 0.9, 1] and the given premise is $A^*(x)$=[1, 0.9, 0.3, 0]. At this time the index is $\Theta = LCM(u,v) = 12$. The dimensions of the extended every vectors $\tilde{A}$, $\tilde{A}^*$, and $\tilde{B}$ are $12 \times 1$ dimension, respectively. The proposed approximate reasoning results are computed by two form, i.e., *P(+1, 0, –1) form* and *P(+1, –1) form*. We obtain the final fuzzy approximate reasoning result based on proposed FMP–LCM; *P(+1, 0, –1) form;* $B^*$=[0, 0.2527, 0.4527, 0.6473, 0.8473, 1], *P(+1, –1) form;* $B^*$=[0, 0.2, 0.4, 0.5946, 0.7946 , 1].

### 3.2. Fuzzy Approximate Reasoning Method For FMT

In this subsection, for the SISO fuzzy system with discrete fuzzy set vectors of different dimensions, we propose a novel FMT–LCM method based on distance measure in the case that differs from dimensions between the antecedent and consequent, i.e., in the case of index $v \neq u$ when element number of the antecedent is $v$ and element number of the consequent is $u$. A novel approximate reasoning method of FMT for SISO fuzzy system with discrete fuzzy set vectors of different dimensions is described as following Stages, which is so called FMT–LCM. A novel FMT-LCM is as follows.

**Stage 1**; Compute the extended fuzzy row vectors. For this let $u, v$ be all real integer index, then the extended fuzzy sets of the antecedent $\bar{B}$, the given premise $B^*$ and consequent $\bar{A}$, i.e., the least common multiple fuzzy vectors $\tilde{\bar{B}}$, $\tilde{B}^* = \{\tilde{B}_l^*\}$, $\tilde{B}_l^* = [\tilde{b}_{1l}^*, \cdots, \tilde{b}_{il}^*, \cdots, \tilde{b}_{ul}^*]$, $l=1,2,\cdots$ and $\tilde{\bar{A}}$ with $\Theta = lcm(u,v)$ are calculated as following, respectively.

$$\tilde{\bar{A}} = \left[\frac{1-a_1}{x_1}, \frac{1-a_2}{x_2}, \cdots, \frac{1-a_r}{x_r}, \cdots, \frac{1-a_{\Theta-1}}{x_{\Theta-1}}, \frac{1-a_\Theta}{x_\Theta}\right] \tag{40}$$

$$\tilde{\tilde{B}} = \left[\frac{1-b_1}{y_1}, \frac{1-b_2}{y_2}, \cdots, \frac{1-b_r}{y_r}, \cdots, \frac{1-b_{\Theta-1}}{y_{\Theta-1}}, \frac{1-b_\Theta}{y_\Theta}\right] \quad (41)$$

$$\tilde{B}_l^* = \left[\frac{b_{1l}^*}{y_1}, \frac{b_{2l}^*}{y_2}, \cdots, \frac{b_{rl}^*}{y_r}, \cdots, \frac{b_{(\Theta-1),l}^*}{y_{\Theta-1}}, \frac{b_{(\Theta),l}^*}{y_\Theta}\right] \quad (42)$$

**Stage 2**; Compute Euclidian distance measure $LCM(\tilde{B}_l^*, \tilde{\tilde{B}})_\Theta$. Where index $l = 1, 2, \cdots$ means the number of the given premise fuzzy set. The distance measure $LCM(\tilde{B}_l^*, \tilde{\tilde{B}})_\Theta (\Theta = lcm(u,v))$ is calculated as follows.

$$LCM(\tilde{B}_l^*, \tilde{\tilde{B}})_\Theta = \left[\frac{1}{\Theta}\sum_{r=1}^{\Theta}[b_{rl}^* - (1-b_r)]^2\right]^{1/2}, \text{ for FMT} \quad (43)$$

**Stage 3**; Compute the sign vectors $\tilde{P}_l$ by the difference $dif_{kl} = b_{kl}^* - (1-b_k), (k = \overline{1, LCM(u,v)}, l = 1, 2, \cdots)$ of the given premise and the antecedent.

$$\tilde{P}_l = [\tilde{P}_{1l}, \tilde{P}_{2l}, \cdots \tilde{P}_{kl}, \cdots], \quad k = \overline{1, LCM(u,v)}, l = 1, 2, \cdots \quad (44)$$

$$\text{P(+1,0,-1) form; } \tilde{P}_{kl} = sign(dif_{kl}) = \begin{cases} +1, & dif_{kl} > 0 \\ 0, & dif_{kl} = 0, \text{ for FMT-LCM} \\ -1, & dif_{kl} < 0 \end{cases} \quad (45)$$

$$\text{P(+1,-1) form; } \tilde{P}_{kl} = sign(dif_{kl}) = \begin{cases} +1, & dif_{kl} \geq 0 \\ -1, & dif_{kl} < 0 \end{cases}, \text{ for FMT-LCM} \quad (46)$$

**Stage 4**; Compute the vectorialized distance measure $\tilde{C}_l$ since distance measure $LCM(\tilde{B}_l^*, \tilde{\tilde{B}})_\Theta$ with an index $\Theta = lcm(u,v)$ is a scalar.

$$\tilde{C}_l = LCM(\tilde{B}_l^*, \tilde{\tilde{B}})_\Theta \times \tilde{P}_l, \Theta = LCM(u,v) \quad (47)$$

**Stage 5**; Obtain the quasi–quasi–approximate reasoning results $\tilde{A}_l^{**}, l = 1, 2, \cdots$ for FMT–LCM.

$$\tilde{A}_l^{**} = \begin{cases} 1 - \tilde{\tilde{A}}_l + \tilde{C}_l, & \text{if Case 6, 7, and 8} \\ \tilde{\tilde{A}}_l + \tilde{C}_l, & \text{if Case 9} \\ s.t.\ \tilde{\tilde{A}}_l + \tilde{C}_l, & \text{if Case 10} \end{cases} \quad (48)$$

**Stage 6**; Select the quasi–approximate reasoning results $A_l^{**}, l = 1, 2, \cdots$ from the quasi–quasi– approximate reasoning results $\tilde{A}_l^{**}$ for indexes $k = \overline{1, lcm(u \cdot v)}$, $l = 1, 2, \cdots$. We will call this FMT–LCM. That is, the reasoning results by the extended dimension $\Theta \times 1$ is transformed into the original dimension $u \times 1$.

$$\tilde{A}_l^{**} \to A_l^{**}, \text{ i.e., } [\tilde{a}_{lk}^{**}]_{\Theta \times 1} \to [a_{lp}^{**}]_{u \times 1} \quad (49)$$

The index $\Theta = LCM(u,v) = u \cdot m_1 = v \cdot m_2$, then $a_{lp}^{**} = \tilde{a}_{l, p \cdot m_1}^{**}$

**Stage 7**; Solve the individual approximate reasoning result $A_l^*$ from the quasi–approximate reasoning results $A_l^{**}$.

$$A_l^* = (A_l^{**} - \eta_l)/(\xi_l - \eta_l), l = 1, 2, \cdots \quad (50)$$

Where, the maximum $\xi_l$ and minimum $\eta_l$ of $A_l^{**}$ is calculated as follows respectively.

$$\xi_l = \max A_l^{**}, \eta_l = \min A_l^{**}. \quad (51)$$

**Stage 8**; For the SISO fuzzy system with discrete fuzzy set vectors of different dimensions, the final approximate reasoning result $A^*$ according to the given premises for FMT–LCM is obtained as follows.

$$A^* = \{A_l^*\}, l = 1, 2, \cdots \quad (52)$$

**Example 3.2** About the proposed FMT−LCM, let us consider for SISO fuzzy system with 6×1 dimension antecedent fuzzy row vector 1−$B(y)$=[1, 0.8, 0.6, 0.3, 0.1, 0] and 4×1 dimension consequent fuzzy row vector 1−$A(x)$=[0, 0.2, 0.6, 1]. When the given premise is $B^*(y)$ =[1, 0.9, 0.8, 0.3, 0.1, 0], let us consider reasoning result of FMT−LCM. The proposed approximate reasoning results are computed by two form, i.e., *P(+1, 0, −1) form* and *P(+1, −1) form*. The index is $\Theta = LCM(u,v) = 12$. The dimensions of the extended every vectors $\tilde{A}$, $\tilde{B}$, and $\tilde{B}^*$ are 12×1 dimension. Therefore the proposed reasoning results for FMT−LCM are computed by two form, i.e., *P(+1, 0, −1) form* and *P(+1, −1) form*. The final results are as follows; *P(+1, 0, −1) form*; $A^*$=[0, 0.2184, 0.5632, 1], *P(+1, −1) form*; $A^*$=[0, 0.2, 0.6, 1].

### 3.3. Reductive Property and Information Loss of the Proposed Method LCM

From proposed method FMP-LCM and FMT-LCM, following theorems with respect to the reductive property and information loss are obtained.

**Theorem 3.1.** *For the SISO fuzzy system, if $A^*(x) = A(x) \subseteq F(X), x \in X$, and is applied FMP−LCM, then the reasoning result $B^*(y) \subseteq F(Y), y \in Y$ is the consequent $B(y) \subseteq F(Y), y \in Y$, thereby the reductive property is completely satisfied. Where $F(X), F(Y)$ are all the fuzzy subsets on the universe of discourse $X, Y$, respectively.*

**Proof.** Since the indexes $u, v$ are all real integer index, then from the extended fuzzy sets of the antecedent $A$, the given premise $A^*$ and consequent $B$, i.e., the least common multiple fuzzy vectors $\tilde{A}$, $\tilde{A}^* = \{\tilde{A}_l^*\}$, $\tilde{A}_l^* = [\tilde{a}_{1l}^*, \cdots, \tilde{a}_{il}^*, \cdots, \tilde{a}_{ul}^*], l=1$, and $\tilde{B}$, new fuzzy approximate reasoning results $B^*$ are calculated as follows, respectively. The difference between the extended every fuzzy vectors $\tilde{A}^*$ and $\tilde{A}$ according to the index $\Theta = LCM(u,v)$, are calculated as follows; $\tilde{A}_l^* - \tilde{A} = \left[\frac{a_{1l}^* - a_1}{x_1}, \cdots, \frac{a_{rl}^* - a_r}{x_r}, \cdots, \frac{a_{\Theta,l}^* - a_\Theta}{x_\Theta}\right]$. From the condition of theorem 3.1, since $a_{rl}^* = a_r, r = \overline{1, \Theta}$, following equation $\tilde{A}_l^* - \tilde{A} = 0$ is satisfied. Therefore, for every index $\Theta$, the distance measure by the least common multiple (LCM), $LCM(\tilde{A}_l^*, \tilde{A})_\Theta = \left[\frac{1}{\Theta}\sum_{r=1}^{\Theta}[a_{rl}^* - a_r]^2\right]^{1/2} = 0$. Then the vectorialized distance measure $\tilde{C}_l = LCM(\tilde{A}_l^*, \tilde{A})_\Theta \times P_l$ is equal to 0, thereby the quasi–quasi–approximate reasoning results is obtained as $\tilde{B}_l^{**} = \tilde{B}_l + \tilde{C}_l = \tilde{B}_l$, and then the quasi–approximate reasoning results is computed as $\tilde{B}_l^{**} \to B_l^{**}$, i.e., $[\tilde{b}_{lk}^{**}]_{\Theta \times 1} \to [b_{lq}^{**}]_{v \times 1}$. From the individual reasoning result $B_l^* = (B_l^{**} - \eta_l)/(\xi_l - \eta_l), l=1$, the final fuzzy reasoning result is obtained as $B^* = B = [b_1, \cdots, b_q, \cdots, b_v]$. Hence This theorem 3.1 is right. Therefore if the given premise discrete fuzzy vector is $A^*(x) = A(x) \subseteq F(X), x \in X$, and the proposed fuzzy approximate reasoning method, i.e., FMP−LCM is applied in the SISO fuzzy system, then the approximate reasoning result $B^*(y) \subseteq F(Y), y \in Y$ of FMP−LCM is obtained to equal to the consequent fuzzy vector $B(y) \subseteq F(Y), y \in Y$ of the fuzzy rule, thereby when $F(X), F(Y)$ are all the fuzzy subsets on the universe of discourse $X, Y$, respectively, the reductive property of the proposed FMP−LCM is completely satisfied. Thus Theorem 3.1 is proved. □

**Theorem 3.2** *For the SISO fuzzy system, if the given premise is $B^*(y) = \overline{B}(y) \subseteq F(Y), y \in Y$, and FMT−LCM is applied in the fuzzy system, then its fuzzy reasoning result $A^*(x) \subseteq F(X), x \in X$ is equal to the consequent $\overline{A}(x) \subseteq F(X), x \in X$, thereby the reductive property is completely satisfied. Where $F(X), F(Y)$ are all the fuzzy subsets on the universe of discourse $X, Y$, respectively.*

Proof is abbreviated. □

**Theorem 3.3** *For the SISO fuzzy system, proposed fuzzy reasoning method FMP−LCM has not information loss.*

**Proof.** Computation process of FMP−LCM is as follows. Where $\tilde{B}_l$ is the least common multiple fuzzy vectors computed by LCM from original antecedent discrete fuzzy vector $\overline{B}$, symbol $\xrightarrow{LCM}$ means transformation by the least common multiple, and $\xrightarrow{LCM^{-1}}$ means reduced transformation of extended vector by LCM. And index $\Theta = LCM(u,v)$ is the least common multiple between dimension number $u$ of fuzzy set $A(x), A^*(x) \subseteq F(X)$ and dimension number $v$ of fuzzy set $B(y) \subseteq F(Y)$. And $\tilde{P}_l = [\tilde{P}_{1l}, \tilde{P}_{2l}, \cdots \tilde{P}_{kl}, \cdots]$, $k = \overline{1, LCM(u,v)}$, $l=1, 2, \cdots$ are sign vectors for addition operation of vectors since the extended distance measure $LCM(\tilde{A}_l^*, \tilde{A})_\Theta$ is scalar.

$$B^* = \begin{cases} [(\begin{bmatrix} \tilde{B}_l + \left[\frac{1}{\Theta}\sum_{r=1}^{\Theta}[a_{rl}^*-a_r]^2\right]^{\frac{1}{2}} \times \tilde{P}_l, \text{ for Case } 1,2,3 \\ 1-\tilde{B}_l + \left[\frac{1}{\Theta}\sum_{r=1}^{\Theta}[a_{rl}^*-a_r]^2\right]^{\frac{1}{2}} \times \tilde{P}_l, \text{ for Case } 4 \\ s.t. \tilde{B}_l + \left[\frac{1}{\Theta}\sum_{r=1}^{\Theta}[a_{rl}^*-a_r]^2\right]^{\frac{1}{2}} \times \tilde{P}_l, \text{ for Case } 5 \end{bmatrix} \xrightarrow{LCM} B_l^{**}) - \min(\begin{bmatrix} \tilde{B}_l + \left[\frac{1}{\Theta}\sum_{r=1}^{\Theta}[a_{rl}^*-a_r]^2\right]^{\frac{1}{2}} \times \tilde{P}_l, \text{ for Case } 1,2,3 \\ 1-\tilde{B}_l + \left[\frac{1}{\Theta}\sum_{r=1}^{\Theta}[a_{rl}^*-a_r]^2\right]^{\frac{1}{2}} \times \tilde{P}_l, \text{ for Case } 4 \\ s.t. \tilde{B}_l + \left[\frac{1}{\Theta}\sum_{r=1}^{\Theta}[a_{rl}^*-a_r]^2\right]^{\frac{1}{2}} \times \tilde{P}_l, \text{ for Case } 5 \end{bmatrix} \xrightarrow{LCM} B_l^{**}))] \\ /[\max(\begin{bmatrix} \tilde{B}_l + \left[\frac{1}{\Theta}\sum_{r=1}^{\Theta}[a_{rl}^*-a_r]^2\right]^{\frac{1}{2}} \times \tilde{P}_l, \text{ for Case } 1,2,3 \\ 1-\tilde{B}_l + \left[\frac{1}{\Theta}\sum_{r=1}^{\Theta}[a_{rl}^*-a_r]^2\right]^{\frac{1}{2}} \times \tilde{P}_l, \text{ for Case } 4 \\ s.t. \tilde{B}_l + \left[\frac{1}{\Theta}\sum_{r=1}^{\Theta}[a_{rl}^*-a_r]^2\right]^{\frac{1}{2}} \times \tilde{P}_l, \text{ for Case } 5 \end{bmatrix} \xrightarrow{LCM} B_l^{**}) - \min(\begin{bmatrix} \tilde{B}_l + \left[\frac{1}{\Theta}\sum_{r=1}^{\Theta}[a_{rl}^*-a_r]^2\right]^{\frac{1}{2}} \times \tilde{P}_l, \text{ for Case } 1,2,3 \\ 1-\tilde{B}_l + \left[\frac{1}{\Theta}\sum_{r=1}^{\Theta}[a_{rl}^*-a_r]^2\right]^{\frac{1}{2}} \times \tilde{P}_l, \text{ for Case } 4 \\ s.t. \tilde{B}_l + \left[\frac{1}{\Theta}\sum_{r=1}^{\Theta}[a_{rl}^*-a_r]^2\right]^{\frac{1}{2}} \times \tilde{P}_l, \text{ for Case } 5 \end{bmatrix} \xrightarrow{LCM} B_l^{**})] \end{cases}$$

$$= \begin{cases} [(\begin{bmatrix} \tilde{B}_l + LCM(\tilde{A}_l^*, \tilde{A})_\Theta \times \tilde{P}_l, \text{ for Case } 1,2,3 \\ 1-\tilde{B}_l + LCM(\tilde{A}_l^*, \tilde{A})_\Theta \times \tilde{P}_l, \text{ for Case } 4 \\ s.t. \tilde{B}_l + LCM(\tilde{A}_l^*, \tilde{A})_\Theta \times \tilde{P}_l, \text{ for Case } 5 \end{bmatrix} \xrightarrow{LCM} B_l^{**}) - \min(\begin{bmatrix} \tilde{B}_l + LCM(\tilde{A}_l^*, \tilde{A})_\Theta \times \tilde{P}_l, \text{ for Case } 1,2,3 \\ 1-\tilde{B}_l + LCM(\tilde{A}_l^*, \tilde{A})_\Theta \times \tilde{P}_l, \text{ for Case } 4 \\ s.t. \tilde{B}_l + LCM(\tilde{A}_l^*, \tilde{A})_\Theta \times \tilde{P}_l, \text{ for Case } 5 \end{bmatrix} \xrightarrow{LCM} B_l^{**}))] \\ /[\max(\begin{bmatrix} \tilde{B}_l + LCM(\tilde{A}_l^*, \tilde{A})_\Theta \times \tilde{P}_l, \text{ for Case } 1,2,3 \\ 1-\tilde{B}_l + LCM(\tilde{A}_l^*, \tilde{A})_\Theta \times \tilde{P}_l, \text{ for Case } 4 \\ s.t. \tilde{B}_l + LCM(\tilde{A}_l^*, \tilde{A})_\Theta \times \tilde{P}_l, \text{ for Case } 5 \end{bmatrix} \xrightarrow{LCM} B_l^{**}) - \min(\begin{bmatrix} \tilde{B}_l + LCM(\tilde{A}_l^*, \tilde{A})_\Theta \times \tilde{P}_l, \text{ for Case } 1,2,3 \\ 1-\tilde{B}_l + LCM(\tilde{A}_l^*, \tilde{A})_\Theta \times \tilde{P}_l, \text{ for Case } 4 \\ s.t. \tilde{B}_l + LCM(\tilde{A}_l^*, \tilde{A})_\Theta \times \tilde{P}_l, \text{ for Case } 5 \end{bmatrix} \xrightarrow{LCM} B_l^{**})] \end{cases}$$

$$= \left\{ [(\tilde{B}_l^{**} \xrightarrow{LCM^{-1}} B_l^{**}) - \min(\tilde{B}_l^{**} \xrightarrow{LCM^{-1}} B_l^{**}))]/[\max(\tilde{B}_l^{**} \xrightarrow{LCM^{-1}} B_l^{**}) - \min(\tilde{B}_l^{**} \xrightarrow{LCM^{-1}} B_l^{**})] \right\}$$

$$= \left\{ (B_l^{**} - \min B_l^{**})/(\max B_l^{**} - \min B_l^{**}) \right\}$$

$$= \left\{ (B_l^{**} - \eta_l)/(\xi_l - \eta_l) \right\}$$

$$= \left\{ B_l^* \right\}, l = 1, 2, \cdots$$

As seen from the formula (25)-(38), since operation processes of Case 1, 2, 3, 4, and 5 have not used nonlinear operators, and accomplished standardization operation $(B_l^{**} - \min B_l^{**})/(\max B_l^{**} - \min B_l^{**})$ is applied, therefore from the above formula, we can easily know that FMP–LCM has not information loss. At this time there is no information loss by $\min B_l^{**}$ and $\max B_l^{**}$. Thus Theorem 3.3 is proved. □

**Theorem 3.4** *For the SISO fuzzy system, proposed fuzzy approximate reasoning method FMT–LCM has not any information loss.*

This proof is similar to Theorem 3.3 as shown the formula (39)-(51), so it is abbreviated.

## 4. Discussion to Controllability of Fuzzy Reasoning Methods

In this section we analyze on [19–22]'s fuzzy reasoning methods based on fuzzy relation and our proposed method based on distance measure with respect to fuzzy control.

### 4.1. Comparison of Fuzzy Reasoning methods based on Fuzzy Relation and Proposed LCM

Let us consider [22]'s method and [32]'s CRI on the basis of the formula (1), (2). According to [19] fuzzy reasoning style is based on the idea converting the fuzzy conditional sentence of 《 *if* $x$ *is* $A$ *then* $y$ *is* $B$ 》 to the fuzzy relation for FMP and FMT. That is,

$$《 x \text{ is } A 》 \to 《 y \text{ is } B 》 = 《 (x, y) \text{ is } R 》 \tag{53}$$

In general fuzzy control, 2 antecedents and 1 consequent are usually used as follows.

$$《 x_1 \text{ is } A_1 》 \text{ and } 《 x_2 \text{ is } A_2 》 \to 《 y \text{ is } B 》 \tag{54}$$

The formula (54) is divided as the formula (55).

$$《 x_1 \text{ is } A_1 》 \to 《 y \text{ is } B 》 = 《 (x_1, y) \text{ is } R 》 \text{ and } 《 x_2 \text{ is } A_2 》 \to 《 y \text{ is } B 》 = 《 (x_2, y) \text{ is } R 》 \tag{55}$$

By expressing like this, we can consider $(x, y)$ to the names of objects, and $R$ in the formula (55) is a fuzzy relation. Denoting fuzzy rule 《 *if $x$ is $A$ then $y$ is $B$* 》 as $A \to B$, then fuzzy relation $A \to B$ is defined as fuzzy implication $\mu_{A \to B}(u, v) = \mu_A(u) \to \mu_B(v)$. At this time, $R$ may be changed according to what implication $A \to B$ is used. They are as follows.

$$R_p = A \times B = \int_{U \times V} \mu_A(u) \mu_B(v)/(u, v) \tag{56}$$

$$R_a = (1 - A) \times B = \int_{U \times V} [1 \wedge (1 - \mu_A(u) + \mu_B(v))]/(u, v) \tag{57}$$

$$R_c = A \times B = \int_{U \times V} \mu_A(u) \wedge \mu_B(v)/(u, v) \tag{58}$$

$$R_m = (A \times B) \bigcup ((1 - A) \times V) = \int_{U \times V} [\mu_A(u) \wedge \mu_B(v) \wedge (1 - \mu_A(u))]/(u, v) \tag{59}$$

Using fuzzy relation $R_c, R_m, R_a$ instead of $A \to B$, conclusion $B^*$ is obtained by FMP–CRI. In the same way, conclusion $A^*$ is obtained by FMT–CRI. There are $R_g$, $R_s$ methods beside fuzzy relation $R_c, R_m, R_a$. In [16] they are as follows.

$$R_s = A \times V \xrightarrow{s} U \times B = \int_{U \times V} \mu_A(u) \xrightarrow{s} \mu_B(v)/(u, v), \quad \mu_A(u) \xrightarrow{s} \mu_B(v) = \begin{cases} 1, & \text{if } \mu_A(u) \leq \mu_B(v) \\ 0, & \text{if } \mu_A(u) > \mu_B(v) \end{cases} \tag{60}$$

$$R_g = A \times V \xrightarrow{g} U \times B = \int_{U \times V} \mu_A(u) \xrightarrow{g} \mu_B(v)/(u, v), \quad \mu_A(u) \xrightarrow{g} \mu_B(v) = \begin{cases} 1, & \text{if } \mu_A(u) \leq \mu_B(v) \\ \mu_B(v), & \text{if } \mu_A(u) > \mu_B(v) \end{cases} \tag{61}$$

By combining fuzzy relation $R_s$ and $R_g$, four fuzzy relations are obtained. And by introducing the accommodation in multi–value logic, several fuzzy relations are obtained.

**4.2. Analysis of Fuzzy Control Reasoning Methods**

In this subsection, we consider about the fuzzy control based on different fuzzy reasoning methods. As shown in [33], the mathematical model of the simple control object is as follows.

$$G(s) = [1/(1 + TS)] \cdot \exp(-\tau s), \quad T = 20(s), \tau = 2(s) \tag{62}$$

Where target value is $r(t) = 40$, the sampling time $t = 1(s)$, error $e(t) = r(t) - y(t)$, change of error $\Delta e(t) = y(t-1) - y(t)$. The increment $\Delta u(k)$ of fuzzy control obtained by fuzzy reasoning is calculated as follows.

$$u(k) = u(k-1) + \rho \cdot \Delta u(k) \tag{63}$$

Where, parameter $\rho$ is amplification coefficient, $k$ discrete time. According to the experimental result, proposed LCM, $R_c$, $R_m$, $R_a$, and so on, can be all applied to fuzzy control. Especially the fuzzy reasoning method $R_c$ presented by E. H. Mamdani in [19] was widely used not only fuzzy control but also pattern recognition, expert system, modeling, predication, system analysis, diagnosis, retrieval system, learning system, and so on. As mentioned in [13–16], the fuzzy relation based reasoning methods $R_{ss}, R_{sg}, R_s, R_{gg}, R_{gs}$, and $R_g$ cannot be applied to fuzzy control. From above consideration, those of reasons can be summarized as follows.

**Theorem 4.1.** *Fuzzy reasoning methods CRI, TIP, and QIP based on the fuzzy relation and/or fuzzy implication and/or t‾norm and/or t‾conorm have some information loss in the fuzzy control reasoning processing.*

**Proof.** Generally the information loss according to [30] can be considered as follows.

(i): To simplify antecedents of fuzzy control rules by using min operator leads to information loss,

(ii): To fire a part of fuzzy control rules with respect to certain threshold leads to information loss,

(iii): To aggregate fuzzy control rules or intermediate conclusions leads to information loss.

In above form (i) when the fuzzy control rule is given as "*if $x_1$ is $A_1$ and $x_2$ is $A_2$ then $y$ is $B$*" then this is equivalent to the following assertion "$A_1 \times A_2 \to B$", fuzzy subset $A_1 \times A_2$ is transformed, for example, by $\min(A_1, A_2)$, and implication $\to$ is computed, according to Łukasiewicz's implication, $\min(A_1 \times A_2) \to B = (1 - \min(A_1 \times A_2) + B) \wedge 1 = (\bullet) \wedge 1$. Since the larger membership degree among $A_1$ and $A_2$ has been lost by $\min(A_1, A_2)$ and the larger membership degree among $(\bullet)$ and 1 has also been lost by $(\bullet) \wedge 1 = \min((\bullet), 1)$, therefore it is obvious that information loss happens always. In above form (ii), widely used method for the simplest fuzzy controller is as follows:

> Step 1; Compute the distance (for example the Euclidian or Hausdorff distance) between the fuzzy input information, "$A_1^*$ and $A_2^*$", and the antecedent of every fuzzy control rules,
> Step 2; Set up certain threshold $\theta$,
> Step 3; Find the fired fuzzy control rules of which the distances of their antecedents from the fuzzy input are smaller than the threshold $\theta$,
> Step 4; Give up all the unfired fuzzy control rules.

From the mathematical viewpoint [17–21, 30], every fuzzy control rule should take part in the procedure of fuzzy reasoning no matter to what extent it acts. Hence the Step 4 leads to information loss. In above form (iii), when the several fuzzy control rules "$if\ x_i\ is\ A_i\ then\ y_i\ is\ B_i, i=\overline{1,n}$" and the fuzzy input $A^*$ are given, then we can first aggregate these fuzzy control rules to construct a super rule, and then infer the fuzzy output $B^*$ according to the Zadeh's compositional rule of inference (CRI). At this time with respect to the aggregation we employ min operator, thereby it would cause information loss. It is evidently clear that the more the fuzzy variable is increase the more the information loss of fuzzy reasoning process is large. The main reason is to nonlinear operators in the fuzzy reasoning process are employed. Since TIP, and QIP are both based on CRI, and employ nonlinear implication operators, and nonlinear t-norms, therefore the information losses in their fuzzy reasoning process are also exist, respectively. Therefore Theorem 4.1 is proved. □

**Theorem 4.2.** *Fuzzy reasoning methods based on the similarity measure expressed by the formula (10)-(13) and (16)-(25) have some information loss in the reasoning processing for fuzzy control.*

**Proof.** This theorem can easily be proved on the basis of [30]. There are a lot of similarity measure (SM) based fuzzy approximate reasoning methods. As mentioned in [30], the fuzzy reasoning methods based on SM depend strongly on the similarity measure and the modification function, and do not completely satisfy the reductive property.

Firstly let us consider the formula (10)-(13). From the formula (10) the larger vale among 1 and $B(y)/SM(A,B)$ has been lost by computation of $B^*(y)=\min\{1,\ B(y)/SM(A,B)\}$. When these fuzzy reasoning methods based on the SM are all applied to the form (i)-(iii) and Step 1- Step 4 of Theorem 4.1, then their information loss is larger. The formula (11)-(13) also are similar to (10). The consideration for this is abbreviated. Therefore fuzzy reasoning methods based on the SM (10)-(13) has some information loss in the fuzzy control reasoning processing.

Next let us consider the formula (16)-(18). Although the formula (16), (17), and (18) have not some information loss, respectively, when they are applied in fuzzy control, information loss is necessarily happened. In other words, when the form (i)-(iii) and Step 1- Step 4 of Theorem 4.1 applied in fuzzy control, then the reasoning results by the formula (16)-(18) have some information losses.

Finally let us consider the formula (19)-(25). we can know that different SM affect differently on the reasoning result. In the formula (19) some information loss occurs by absolute value $|\mu_A(u)-\mu_B(v)|$ operation and $\max_{u\in U}(\bullet)$ operation. And in the formula (20) and (24), the result of operation by $\mu_A(u)\cdot\mu_B(v)$ is less than $\mu_A(u), \mu_B(v)$, respectively, thus information loss happens always. And in the formula (21), (22), and (23), the larger membership degree among $\mu_A(u)$ and $\mu_B(v)$ has been lost by $\min(\mu_A(u),\mu_B(v))$, whereas, the smaller membership degree among $\mu_A(u)$ and $\mu_B(v)$ has been lost by $\max(\mu_A(u),\mu_B(v))$. The formula (25) has some information loss by form (i)-(iii) and Step 1- Step 4 of Theorem 4.1. Therefore it is obvious that information loss happens in the formula (19)-(25). Thus Theorem 4.2 is proved.□

**Theorem 4.3.** *Fuzzy reasoning methods based on the fuzzy relation $R_s$ and $R_g$ presented by the formula (60), (61) do not satisfy the convergence of the fuzzy control.*

**Proof.** Let $u_0 \in U$ be crisp input information, $\mu_{A^*}(u_0)$ membership function fuzzificated by crisp value $u_0 \in U$. For every crisp information value $u_0 \in U$, the fuzzy reasoning results by $R_s$ and $R_g$ are always obtained as 1 or 0 for $R_s$, and 1 or $\mu_B(v)$ for $R_g$, respectively, that is, according to CRI [32], conclusions $\mu_{B^*}(v)$ of fuzzy reasoning method $R_s$ and $R_g$ are obtained as follows, respectively.

$$\mu_{B^*}(v)=\vee_u\{\mu_{A^*}(u_0)\wedge[\mu_A(u)\underset{s}{\to}\mu_B(v)]\}=\vee_u\{\mu_{A^*}(u_0)\wedge\mu_A(u)\}\underset{s}{\to}\mu_B(v)=h\underset{s}{\to}\mu_B(v)=\begin{cases}1,& h\le\mu_B(v)\\0,& h>\mu_B(v)\end{cases} \quad (64)$$

$$\mu_{B^*}(v)=\vee_u\{\mu_{A^*}(u_0)\wedge[\mu_A(u)\underset{g}{\to}\mu_B(v)]\}=\vee_u\{\mu_{A^*}(u_0)\wedge\mu_A(u)\}\underset{g}{\to}\mu_B(v)=h\underset{g}{\to}\mu_B(v)=\begin{cases}1,& h\le\mu_B(v)\\\mu_B(v),& h>\mu_B(v)\end{cases} \quad (65)$$

, where $h=\vee_u\{\mu_{A^*}(u_0)\wedge\mu_A(u)\}$ is a degree of matching of the fuzzy control rule. From the formula (64) and (65) we can know that when input information $\mu_{A^*}(u_0)$ is changed according to $u_0$ then reasoning result $\mu_{B^*}(v)$ is not changed and fixed as crisp value 1 and 0, and fuzzy set $\mu_B(v)$, therefore the convergence of the fuzzy control cannot be guaranteed. As mentioned in [16,17] these are logical contradiction between the reductive property and the practical problem (e.g. fuzzy control) Thus Theorem 4.3 is proved.□

**Theorem 4.4.** *Fuzzy reasoning methods based on the fuzzy relation $R_{ss}, R_{sg}, R_s, R_{gg}, R_{gs}$, and $R_g$ presented by the formula (60), (61), and their combinations do not satisfy the convergence of the fuzzy control.*

**Proof**. From Theorem 4.3 we can easily know the convergence of the fuzzy control. When the fuzzy relation $R_{ss}, R_{sg}, R_s, R_{gg}, R_{gs}$, and $R_g$ are applied to fuzzy control, for different input information different reasoning results are not obtained, but same ones are also calculated, respectively. That is, those have not their convergence. Thisese can be illustratively proved by the extension of formula (64) and (65), respectively.□

**Theorem 4.5.** *Fuzzy reasoning methods based on the fuzzy relation $R_c, R_m, R_p$ and $R_a$ presented by the formula (56)- (59) do satisfy the convergence of the fuzzy control.*

**Proof.** For different input information, different reasoning results are obtained by the fuzzy relation $R_c, R_m, R_p$ and $R_a$. First, let us consider Mamdani's fuzzy relation $R_c$. For crisp information value $u_0 \in U$, let $h_i = \bigvee_u \{\mu_{A^*}(u_0) \wedge \mu_{A_i}(u)\}$ be degree of matching of the *ith* rule, and $n$ number of rules, then the individual fuzzy reasoning results by $R_c$ are as follows.

$$\mu_{B_i^*}(v) = \bigvee_u \{\mu_{A^*}(u_0) \wedge [\mu_{A_i}(u) \underset{Rc}{\to} \mu_{B_i}(v)]\} = \bigvee_u \{\mu_{A^*}(u_0) \wedge \mu_{A_i}(u)\} \underset{Rc}{\to} \mu_{B_i}(v) = h_i \wedge \mu_{B_i}(v) \quad (66)$$

The final fuzzy reasoning result $\mu_{B_i^*}(v)$ by fuzzy relation $R_c$ and defuzzificated crisp value $v_0$ by center of gravity are calculated as follows, respectively.

$$\mu_{B_i^*}(v) = \bigcup_{i=1}^n \{h_i \wedge \mu_{B_i}(v)\} \quad (67)$$

$$v_0 = \bigcup_{i=1}^n [h_i \wedge \mu_{B_i}(v)] / \bigcup_{i=1}^n h_i \quad (68)$$

From formula (67) and (68), we can see that when degree of matching $h_i$ is changed then $\mu_{B^*}(v)$ and $v_0$ are also changed according to $h_i$, therefore the fuzzy reasoning by Mamdani's $R_c$ ([13]) has the convergence of the fuzzy control.

Next the fuzzy reasoning results by $R_p, R_a, R_m$ are as follows, respectively.

$$\mu_{B_i^*}(v) = \bigvee_u \{\mu_{A^*}(u_0) \wedge [\mu_{A_i}(u) \underset{Rp}{\to} \mu_{B_i}(v)]\} = \bigvee_u \{\mu_{A^*}(u_0) \wedge \mu_{A_i}(u)\} \underset{Rp}{\to} \mu_{B_i}(v) = h_i \times \mu_{B_i}(v), \text{ for } R_p \quad (69)$$

$$\mu_{B_i^*}(v) = \bigvee_u \{\mu_{A^*}(u_0) \wedge [\mu_{A_i}(u) \underset{Ra}{\to} \mu_{B_i}(v)]\} = \bigvee_u \{\mu_{A^*}(u_0) \wedge \mu_{A_i}(u)\} \underset{Ra}{\to} \mu_{B_i}(v) = 1 \wedge [(1 - h_i) + \mu_{B_i}(v)], \text{ for } R_a \quad (70)$$

$$\mu_{B_i^*}(v) = \bigvee_u \{\mu_{A^*}(u_0) \wedge [\mu_{A_i}(u) \underset{Rm}{\to} \mu_{B_i}(v)]\} = \bigvee_u \{\mu_{A^*}(u_0) \wedge \mu_{A_i}(u)\} \underset{Rm}{\to} \mu_{B_i}(v) = h_i \wedge \mu_{B_i}(v) \wedge (1 - h_i)$$
$$= \begin{cases} h_i \wedge \mu_{B_i}(v), & \text{if } h_i \le 0.5 \\ (1 - h_i) \wedge \mu_{B_i}(v), & \text{if } h_i > 0.5 \end{cases}; \quad \text{for } R_m \quad (71)$$

The final fuzzy reasoning results $\mu_{B_i^*}(v)$ by fuzzy relation $R_p, R_a, R_m$ are calculated as follows, respectively.

$$\mu_{B_i^*}(v) = \bigcup_{i=1}^n \{h_i \times \mu_{B_i}(v)\}, \text{ for } R_p \quad (72)$$

$$\mu_{B_i^*}(v) = \bigcup_{i=1}^n \{1 \wedge [(1 - h_i) + \mu_{B_i}(v)]\}, \text{ for } R_a \quad (73)$$

$$\mu_{B_i^*}(v) = \bigcup_{i=1}^n \begin{cases} h_i \wedge \mu_{B_i}(v), & \text{if } h_i \le 0.5 \\ (1 - h_i) \wedge \mu_{B_i}(v), & \text{if } h_i > 0.5 \end{cases}, \text{ for } R_m \quad (74)$$

From formula (72)-(74), we can know that when degree of matching $h_i$ is changed then $\mu_{B^*}(v)$ and $v_0$ are also changed according to $h_i$, with respect to $R_p, R_a, R_m$, thus the fuzzy reasoning results by fuzzy relations $R_p, R_a, R_m$ have also the convergence of the fuzzy control. Therefore Theorem 4.5 is proved. □

**Theorem 4.6.** *Our proposed fuzzy reasoning method LCM presented by the theorem 3.1 and 3.2 in this paper has the*

*convergence of the fuzzy control and has not information loss with respect to FMP and FMT.*

**Proof.** For different input information, different reasoning results are obtained by proposed method LCM. That is, for every crisp input information $u_0 \in U$, extended Euclidian distance measures $LCM(\mu_{A^*}(u_0), \mu_{Ai}(u))$ between the given premise $\mu_{A^*}(u_0)$ and antecedent $\mu_{Ai}(u)$ of *ith* fuzzy control rule are not equal to. That is,

$$LCM(\mu_{A^*}(u_0), \mu_{A_1}(u)) \neq \cdots \neq LCM(\mu_{A^*}(u_0), \mu_{A_i}(u)) \neq \cdots \neq LCM(\mu_{A^*}(u_0), \mu_{A_n}(u)) \tag{75}$$

From the proposed method, $v_0 \in V$ satisfying following formula must be obtained.

$$LCM(\mu_B(v_0), \mu_{B_i}(v)) = LCM(\mu_{A^*}(u_0), \mu_{Ai}(u)) \tag{76}$$

Our aim is to find a defuzzificated value, the crisp one $v_0 \in V$ satisfying the formula (76). Then distance measures $LCM(\mu_B(v_0), \mu_{B_i}(v))$ between the fuzzy reasoning conclusion and the consequent fuzzy set of *ith* fuzzy control rule are obtained as follows.

$$LCM(\mu_B(v_0), \mu_{B_1}(v)) \neq \cdots \neq LCM(\mu_B(v_0), \mu_{B_i}(v)) \neq \cdots \neq LCM(\mu_B(v_0), \mu_{B_n}(v)) \tag{77}$$

Where $v_0 \in V$ is a defuzzificated value of the fuzzy reasoning result. Thus the fuzzy reasoning results $\mu_B(v_0)$ are changed to different values $v_0 \in V$ with respect to different $u_0 \in U$. In other words for small input information, corresponding small reasoning results are obtained, whereas, for big input information, corresponding big reasoning results are obtained, which means that fuzzy reasoning method has in itself convergence. Therefore our proposed method does satisfy the reductive property and can be applied to the fuzzy control and so on.

Next proposed method LCM has not information loss. Its important reason is as follows. Let quasi–fuzzy reasoning result be $\tilde{B}_i$, then following formula is satisfied.

$$\mu_{\tilde{B}_i}(v) = \mu_{B_i}(v) + LCM(\mu_{A^*}(u_0), \mu_{A_i}(u)) \times P_i \tag{78}$$

Thus the fuzzy reasoning result of *ith* fuzzy control rule is calculated as follows.

$$\mu_{B_i^*}(v) = (\mu_{\tilde{B}_i}(v) - \eta_i)/(\xi_i - \eta_i) \tag{79}$$

Where $\xi_i = \max_{1 \leq k \leq r} \tilde{B}_i$ and $\eta_i = \min_{1 \leq k \leq r} \tilde{B}_i$ are maximum and minimum of the quasi–fuzzy reasoning result, respectively. By these operations the information losses are overcame. Since the standardizationization operation is used in formula (79), thereby proposed method LCM has not any information loss. Final crisp result $v_0 \in V$ mentioned in formula (76) and (77) is calculated as follows.

$$v_0 = (\sum_{i=1}^{n} B_i^*(v_i) \times v_i)/(\sum_{i=1}^{n} B_i^*(v_i)) \tag{80}$$

Where $v_0, v_i \in V$. From formula (80) we can know that the crisp value $v_0 \in V$ has entirely not the information loss. Thus the proof of this Theorem 4.6 is completed.□

From the theorem 4.1–4.6, we know that our proposed method LCM and $R_c, R_p, R_a, R_m$ all have the convergence of the fuzzy control, i.e., fuzzy controllability. Whereas $R_{sg}, R_s, R_{gg}, R_{gs}, R_g$ have not all fuzzy controllability.

## 5. Comparisons of CRI, TIP, QIP, AARS, and Proposed Method LCM

In this section we discuss on the reductive property of our's and CRI, TIP, QIP, and AARS methods and consider on experiment results of the fuzzy controllability.

### 5.1. Checking of the Proposed Method

**Experiment 5.1** We have done experiment by MATLAB 2017. For the SISO fuzzy system with discrete fuzzy set vectors of different dimensions in Class 1, let us consider an approximate fuzzy reasoning based on 5×1 dimension antecedent fuzzy row vector $A(x)$=[1, 0.3, 0, 0, 0] and 7×1 dimension consequent fuzzy row vector $B(y)$=[0, 0, 0, 0, 0, 0.3, 1] with respect to FMP–LCM. The given premises are $A_1^*(x) = A(x)$ =[1, 0.3, 0, 0, 0], $A_2^*(x) = very\ A(x)$ =[1, 0.09, 0, 0, 0], $A_3^*(x) = more\ or\ less\ A(x)$ =[1, 0.55, 0, 0, 0], $A_4^*(x) = not\ A(x)$ =[0, 0.7, 1, 1, 1], respectively. The proposed approximate reasoning results are computed by two form, i.e., *P(+1,0,–1) form* and *P(+1,–1) form*. The computational fuzzy approximate reasoning result by MATLB experiment is shown in Table 2.

**Table 2.** FMP–LCM Reasoning Results and Reductive Properties in Class 1

| FMP–LCM Premise $A^*(x)$ | | FMP–LCM Reasoning Results and Reductive Property | | | |
|---|---|---|---|---|---|
| | | | Reasoning Results $B^*(y)$ | | RPCF |
| $A_1^*$ | [1, 0.3, 0, 0, 0] | $B_1^*$ | P(+1,0,–1) form | [0, 0, 0, 0, 0, 0.3, 1] | 100 (%) |
| | | | P(+1,–1) form | [0, 0, 0, 0, 0, 0.3, 1] | 100 (%) |
| $A_2^*$ | [1, 0.09, 0, 0, 0] | $B_2^*$ | P(+1,0,–1) form | [0.072,0,0,0,0.072,0.35,1] | 94.24% |
| | | | P(+1,–1) form | [0.13,0,0,0,0.13,0.39,1] | 91.85% |
| $A_3^*$ | [1, 0.55, 0, 0, 0] | $B_3^*$ | P(+1,0,–1) form | [0,0.091,0.091,0.091,0,0.3,1] | 92.56% |
| | | | P(+1,–1) form | [0, 0, 0, 0, 0, 0.3, 1] | 96.46 (%) |
| $A_4^*$ | [0, 0.7, 1, 1, 1] | $B_4^*$ | P(+1,0,–1) form | [0, 0, 1, 1, 1, 0.83, 0.43] | 63.53 (%) |
| | | | P(+1,–1) form | [0, 0, 1, 1, 1, 0.83, 0.43] | 63.53 (%) |
| RPCF | | | FMP–LCM−P(+1,0,–1) form | | 87.58% |
| | | | FMP–LCM−P(+1,–1) form | | 87.96% |

**Experiment 5.2** For the SISO fuzzy system with discrete fuzzy set vectors of different dimensions in Class 2, let us consider an approximate fuzzy reasoning based on 5×1 dimension antecedent fuzzy row vector $A(x)$=[1, 0.3, 0, 0, 0] and 7×1 dimension consequent fuzzy row vector $B(y)$=[0, 0, 0, 0, 0, 0.3, 1] with respect to FMP–LCM. The given premises are $A_1^*(x) = A(x)$ =[1, 0.3, 0, 0, 0], $A_2^*(x) = very\ A(x)$ =[1, 0.09, 0, 0, 0], $A_3^*(x) = more\ or\ less\ A(x)$ =[1, 0.55, 0, 0, 0], $A_4^*(x) = s.t.A(x)$ =[1, 0.2, 0, 0, 0], respectively. The proposed fuzzy approximate reasoning results are computed by two form, i.e., *P(+1,0,–1) form* and *P(+1,–1) form*. The computational fuzzy approximate reasoning result by MATLB is shown in Table 3.

**Table 3.** FMP–LCM Reasoning Results and Reductive Properties in Class 2

| FMP–LCM Premise $A^*(x)$ | | FMP–LCM Reasoning Results and Reductive Property | | | |
|---|---|---|---|---|---|
| | | | Reasoning Results $B^*(y)$ | | RPCF |
| $A_1^*$ | [1, 0.3, 0, 0, 0] | $B_1^*$ | P(+1,0,–1) form | [0, 0, 0, 0, 0, 0.3, 1] | 100 % |
| | | | P(+1,–1) form | [0, 0, 0, 0, 0, 0.3, 1] | 100 % |
| $A_2^*$ | [1, 0.09, 0, 0, 0] | $B_2^*$ | P(+1,0,–1) form | [0.072, 0, 0, 0, 0.072, 0.35, 1] | 94.24 % |
| | | | P(+1,–1) form | [0.13, 0, 0, 0, 0.13, 0.39, 1] | 91.85 % |
| $A_3^*$ | [1, 0.55, 0, 0, 0] | $B_3^*$ | P(+1,0,–1) form | [0, 0.091, 0.091, 0.091, 0, 0.3, 1] | 92.56 % |
| | | | P(+1,–1) form | [0, 0, 0, 0, 0, 0.3, 1] | 96.46 % |
| $A_4^*$ | [1, 0.2, 0, 0, 0] | $B_4^*$ | P(+1,0,–1) form | [0.035, 0, 0, 0, 0.035, 0.23, 1] | 98.58 % |
| | | | P(+1,–1) form | [0.068, 0, 0, 0, 0.068, 0.25, 1] | 97.26 % |
| RPCF | | | FMP–LCM−P(+1,0,–1) form | | 96.35 % |
| | | | FMP–LCM−P(+1,–1) form | | 96.39 % |

**Experiment 5.3** For the SISO fuzzy system with discrete fuzzy set vectors of different dimensions in Class 1 and Class 2, experiment results of LCM are shown in Table 4.

**Table 4.** Comprehension of our proposed FMP–LCM in Class 1 and Class 2

| RPCF | Class 1 | Class 2 | Average |
|---|---|---|---|
| FMT–LCM−P(+1,0,–1) form | 87.58% | 96.35 % | 91.97 % |
| FMT–LCM−P(+1,–1) form | 87.96 % | 96.39 % | 92.18% |

**Experiment 5.4** For the SISO fuzzy system with discrete fuzzy set vectors of different dimensions, let us consider an approximate fuzzy reasoning based on 7×1 dimension antecedent fuzzy row vector $B(y)$=[0, 0, 0, 0, 0, 0.3, 1] and 5×1 dimension consequent fuzzy row vector $A(x)$=[1, 0.3, 0, 0, 0] for FMT–LCM. It is shown in Table 5.

**Experiment 5.5** For the SISO fuzzy system with discrete fuzzy set vectors with different dimensions in Class 1 and Class 2, let us consider the comprehensive reductive property of our proposed LCM method. It is shown in Table 6.

**Table 5.** Reductive Property of FMT–LCM in Class 1

| FMT–LCM−P(+1,0,–1) form | 88.88 (%) |
|---|---|
| FMT–LCM−P(+1,–1) form | 89.01 (%) |

**Table 6.** Comprehension of LCM in Class 1 and Class 2

| Class 1 and 2 | LCM−P(+1,0,–1) | LCM−P(+1,–1) |
|---|---|---|
| RPCF–average | 90.473 % | 90.65 % |

From the examples of Table 2 to Table 6, we can know that the reductive property of the proposed LCM method in the SISO fuzzy system with discrete fuzzy set vectors of different dimensions between the antecedent and consequent of the fuzzy rule, about the comprehension of Class 1 and 2, for *LCM−P(+1,0,–1)form* and *LCM−P(+1,–1)form*, is 90.063%.

## 5.2. Property of Proposed Method LCM

**Property 5.1** Proposed method LCM deals with normal fuzzy sets as the paper [12, 13], whereas, deals not with non–normal fuzzy sets, because non–normal fuzzy sets can be transferred into normal fuzzy sets. In the real world, a lot of engineers and designers actually use normal fuzzy sets when fuzzy sets are applied in the fuzzy systems.

**Property 5.2** the proposed method LCM contains previous method DMM [12, 13], that is, FMP–LCM, FMT–LCM contain FMP–DM, FMT–DM, respectively.

**Property 5.3** The reductive property of $P(+1, -1)$ form in the proposed method LCM is better than $P(+1, 0, -1)$ form with respect to FMP and FMT, and, Class 1 and Class 2.

**Property 5.4** CRI, TIP, QIP and AARS use nonlinear operators such as Max, Min, implications, t–norms, and t–conorms, whereas, the proposed method LCM uses linear operators.

**Property 5.5** The proposed method LCM has not information loss in the fuzzy approximate reasoning process as the previous method DMM [12, 13], because the formula (37) for FMP and (50) for FMT do use respectively the standardization operator.

### 5.3. Comprehensive Comparisons of CRI, TIP, QIP, AARS and proposed LCM in Class 1

The reductive properties of CRI, TIP, QIP, AARS, and proposed LCM in Class 1 are shown in Table 7. Through the MATLB experiments we can obtain the following propositions in Class 1.

**Table 7.** Reductive Properties of CRI, TIP, QIP, AARS, and proposed LCM in Class 1

| No | In Class 1 Reasoning Method | | $RPCF-FMP$ | $RPCF-FMT$ | $RPCF_{FR}$ | Average |
|----|---|---|---|---|---|---|
| 1 | Proposed | $P(+1,0,-1)$ form | 87.58% | 88.88% | 88.23% | 88.36% |
| 2 | LCM | $P(+1,-1)$ form | 87.96 % | 89.01% | 88.49% | |
| 3 | | Gödel; G | 94.33 % | 61.81 % | 78.07% | |
| 4 | CRI | Gougen; Go | 94.61 % | 61.81 % | 78.21% | 76.26% |
| 5 | (1975) | Łukasiewicz; L | 90.18% | 61.81 % | 76.00 % | |
| 6 | | $R_0$ | 83.71 % | 61.81% | 72.76% | |
| 7 | | Gödel; G | 94.33% | 43.99% | 69.16% | |
| 8 | TIP | Gougen; Go | 94.61 % | 44.69 % | 69.65% | 67.61% |
| 9 | (1999) | Łukasiewicz | 90.18% | 44.69% | 67.44% | |
| 10 | | $R_0$ | 83.71 % | 44.69% | 64.20% | |
| 11 | QIP | Łukasiewicz | 77.29% | 42.45% | 59.87% | |
| 12 | (2015– | Gödel; G | 77.29% | 42.45% | 59.87% | 59.59% |
| 13 | 2018) | $R_0$ | 76.22 % | 41.26% | 58.74% | |
| 14 | | Gougen; Go | 77.29% | 42.45% | 59.87% | |
| 15 | AARS | *reduction form* | 76.11 % | 39.19 % | 57.65% | 57.22% |
| 16 | (1990) | *more or less form* | 76.45 % | 37.10 % | 56.78% | |

**Proposition 5.1** For FMP, the reductive property of CRI–Gougen and TIP–Gougen among 16 individual fuzzy approximate reasoning methods dealt with in this paper is best high, whereas, QIP–Gougen best low, with respect to Class 1 in SISO fuzzy system with discrete fuzzy set vectors of different dimensions.

**Proposition 5.2** For FMT, the reductive property of LCM–$P(+1,0,-1)$ form and LCM–$P(+1,-1)$ form among 16 individual fuzzy approximate reasoning methods is high, whereas, AARS–*reduction form* and AARS–*more or less form* best low, in Class 1 for SISO fuzzy system with discrete fuzzy set vectors of different dimensions.

**Proposition 5.3** For average of FMP and FMT, the reductive property of LCM among 5 fuzzy approximate reasoning methods is best high, and then CRI, TIP, QIP whereas, AARS best low, in Class 1 for SISO fuzzy system with discrete fuzzy set vectors of different dimensions.

### 5.4. Comparisons of CRI, TIP, QIP, AARS and Proposed Method LCM in Class 2

In this subsection, we compare and analyze about CRI, TIP, QIP, AARS and proposed LCM method in Class 2. The reductive properties of five fuzzy reasoning methods for Class 2 are shown in Table 8.

**Table 8.** Reductive Properties of CRI, TIP, QIP, AARS and proposed LCM in Class 2

| No | In Class 1 | $RPCF-FMP$ | $RPCF-FMT$ | $RPCF_{FR}$ | Average |
|----|---|---|---|---|---|

| | | | | | | |
|---|---|---|---|---|---|---|
| 1 | Proposed | $P(+1,0,-1)$ form | 96.35 % | 89.08% | 92.72% | 92.77% |
| 2 | LCM | $P(+1,-1)$ form | 96.39 % | 89.22% | 92.81% | |
| 3 | | Gödel; G | 98.61% | 61.31% | 79.96% | |
| 4 | CRI | Gougen; Go | 98.89% | 61.31% | 80.1 % | 78.15% |
| 5 | (1975) | Łukasiewicz; L | 94.47% | 61.31% | 77.89% | |
| 6 | | $R_0$ | 87.99% | 61.31% | 74.65% | |
| 7 | | Gödel; G | 98.61% | 39.09% | 68.85% | |
| 8 | TIP | Gougen; Go | 98.89% | 43.02% | 70.96% | 68.51% |
| 9 | (1999) | Łukasiewicz | 94.47% | 44.19% | 69.33% | |
| 10 | | $R_0$ | 87.99% | 41.79% | 64.89% | |
| 11 | QIP | Łukasiewicz | 98.01% | 41.95% | 69.98% | |
| 12 | (2015– | Gödel; G | 98.01% | 41.95% | 69.98% | 69.83% |
| 13 | 2018) | $R_0$ | 98.01% | 40.76% | 69.38% | |
| 14 | | Gougen; Go | 98.01% | 41.95 % | 69.98% | |
| 15 | AARS | reduction form | 97.98% | 36.56% | 62.27% | 65.12% |
| 16 | (1990) | more or less form | 97.35% | 38.59% | 67.97% | |

From the MATLB experiments we can obtain the following propositions in Class 2.

**Proposition 5.4** For FMP, the reductive property of CRI−Gougen, TIP−Gougen and TIP−Gödel among the individual 16 fuzzy approximate reasoning methods is best high, whereas, AARS−*reduction form* and AARS−*more or less form* best low, in Class 2 for SISO fuzzy system with discrete fuzzy set vectors of different dimensions.

**Proposition 5.5** For FMT, the reductive property of LCM−*P(+1,0,−1) form* and LCM−*P(+1,−1) form* among the individual 16 fuzzy approximate reasoning methods is best high, whereas, AARS−*reduction form* and AARS−*more or less form* best low, in Class 2 for SISO fuzzy system with discrete fuzzy set vectors of different dimensions.

**Proposition 5.6** For average of FMP and FMT, the reductive property of proposed LCM among 5 fuzzy approximate reasoning methods is best high, and then CRI, TIP, QIP whereas, AARS best low, in Class 2 for SISO fuzzy system with discrete fuzzy set vectors of different dimensions.

### 5.5. Comprehensive Analysis of Class 1 and Class 2

The total reductive properties of the 5 fuzzy reasoning methods in Class 1 and Class 2 are comprehensively shown in Table 9. From Table 9 we can obtain the following propositions in Class 1 and Class 2.

**Table 9.** The comprehensive reductive properties of the 5 fuzzy reasoning methods for Class 1 and Class 2

| Fuzzy Reasoning | FMP | FMT | Average |
|---|---|---|---|
| Proposed LCM | 92.07% | 89.05% | 90.56% |
| CRI | 92.85% | 61.56% | 77.21% |
| TIP | 92.85% | 43.27% | 68.06% |
| QIP | 87.52% | 41.90% | 64.71% |
| AARS | 86.97% | 37.86% | 61.17% |

The comparison experiment of the fuzzy approximate reasoning computing times for CRI, TIP, QIP, AARS, and proposed method LCM has done. The experiment was accomplished via 6[th] of test, these average values are shown in Table 10.

**Table 10.** Comparison of the fuzzy approximate reasoning computing times by MATLAB

| Reasoning Method | AARS [27,28] | Proposed LCM | CRI [38] | TIP [10,16,32] | QIP [17,40] |
|---|---|---|---|---|---|
| Computing Time (ms) | 233 | 250 | 255 | 260 | 282 |

**Proposition 5.7** AARS's computational time is best shorted, and then our proposed LCM method, CRI, TIP, and then QIP's one is longest.

**Proposition 5.8** For FMP, in Class 1 and Class 2, the reductive property of CRI and TIP among 5 fuzzy approximate reasoning methods is best high, whereas, AARS best low, for SISO fuzzy system with discrete fuzzy set vectors of different dimensions.

**Proposition 5.9** For FMT, Class 1 and Class 2, the reductive property of proposed LCM among 5 fuzzy approximate reasoning methods is best high, whereas, AARS best low, for SISO fuzzy system with discrete fuzzy set vectors of different dimensions.

**Proposition 5.10** For FMP and FMT, Class 1 and Class 2, among 5 fuzzy approximate reasoning methods, the reductive property of LCM is best high, and then CRI, TIP, whereas, AARS best low, for SISO fuzzy system with discrete fuzzy set vectors of different dimensions. And the proposed method LCM is well in accordance with human thinking than other reasoning methods.

Comprehensively among the 5 fuzzy reasoning methods for Class 1 and Class 2, our proposed LCM method is highest with respect to the reductive property.

### 5.6. Fuzzy Controllability of Several Fuzzy Reasoning Methods

The several fuzzy reasoning results calculated for FMP and FMT presented in [13,16,17,32] and proposed LCM are shown in Table 11, 12, 13 and 14, respectively.

**Table 11.** FMP reasoning results by different fuzzy relations ([See 16,17])

| Method | $A$ | very $A$ | more or less $A$ | not $A$ |
|---|---|---|---|---|
| $R_m$ [32] | $\frac{1}{2} \vee \mu_B$ | $\frac{1}{2}(\sqrt{5}-1) \vee \mu_B$ | $\frac{1}{2}(3-\sqrt{5}) \vee \mu_B$ | 1 |
| $R_a$ [32] | $\frac{1}{2}(1+\mu_B)$ | $\frac{1}{2}(3+2\mu_B-\sqrt{5-4\mu_B})$ | $\frac{1}{2}\sqrt{5+4\mu_B}-1$ | 1 |
| $R_c$ [13] | $\mu_B$ | $\mu_B$ | $\mu_B$ | $\frac{1}{2} \wedge \mu_B$ |
| $R_s$, $R_g$ [17] | $\mu_B$ | $\mu_B^2$ | $\mu_B^{0.5}$ | 1 |
| $R_{sg}$, $R_{gs}$, $R_{gg}$ [17] | $\mu_B$ | $\mu_B^2$ | $\mu_B^{0.5}$ | $1-\mu_B$ |

**Table 12.** FMT reasoning results by different fuzzy relation ([See 16,17])

| Method | not $B$ | not very $B$ | not more or less $B$ | $B$ |
|---|---|---|---|---|
| $R_m$ [32] | $0.5 \vee (1-\mu_A)$ | $(\overline{\mu_A} \vee \frac{1}{2}(\sqrt{5}-1)) \vee \mu_A$ | $\frac{1}{2}(3-\sqrt{5}) \vee (1-\mu_A)$ | $\mu_A \vee (1-\mu_A)$ |
| $R_a$ [32] | $1-\frac{1}{2}\mu_A$ | $\frac{1}{2}(1-2\mu_A+\sqrt{1+4\mu_A})$ | $\frac{1}{2}(3-\sqrt{1+4\mu_A})$ | 1 |
| $R_c$ [13] | $0.5 \vee \mu_A$ | $\frac{1}{2}(\sqrt{5}-1) \wedge \mu_A$ | $\frac{1}{2}(3-\sqrt{5}) \wedge \mu_A$ | $\mu_A$ |
| $R_s$ [17] | $1-\mu_A$ | $1-\mu_A^2$ | $1-\mu_A^{0.5}$ | 1 |
| $R_g$ [17] | $0.5 \vee (1-\mu_A)$ | $\frac{1}{2}(\sqrt{5}-1) \vee (1-\mu_A^2)$ | $\frac{1}{2}(3-\sqrt{5}) \vee (1-\mu_A^{0.5})$ | 1 |
| $R_{sg}$ [17] | $1-\mu_A$ | $1-\mu_A^2$ | $1-\mu_A^{0.5}$ | $0.5 \vee \mu_A$ |
| $R_{gg}$ [17] | $0.5 \vee (1-\mu_A)$ | $\frac{1}{2}(\sqrt{5}-1) \vee (1-\mu_A^2)$ | $\frac{1}{2}(3-\sqrt{5}) \vee (1-\mu_A^{0.5})$ | $0.5 \vee \mu_A$ |
| $R_{gs}$ [17] | $0.5 \vee (1-\mu_A)$ | $\frac{1}{2}(\sqrt{5}-1) \vee (1-\mu_A^2)$ | $\frac{1}{2}(\sqrt{5}-1) \vee (1-\mu_A^2)$ | $\mu_A$ |
| $R_{ss}$ [17] | $1-\mu_A$ | $1-\mu_A^2$ | $1-\mu_A^{0.5}$ | $\mu_A$ |

**Table 13.** Proposed FMP–LCM and FMT–LCM Reductive Property for [17]'s problem

| Antecedent; $A$=[0, 0.25, 0.5, 0.75, 1], Consequent; $B$=[0, 0.25, 0.5, 0.75, 1] | | | |
|---|---|---|---|
| The given premise; $A^*(x)$ | | Conclusion; $B^*(y)=?$ | RPCF |
| $A = [large]$ | $A^* = [0, 0.250, 0.500, 0.750, 1]$ | $B^* = [0, 0.250, 0.500, 0.750, 1]$ | 100 % |
| very $A = [large]^2$ | $A^* = [0, 0.063, 0.250, 0.563, 1]$ | $B^* = [0, 0.087, 0.337, 0.587, 1]$ | 97.28 % |
| more or less $A = [large]^{\frac{1}{2}}$ | $A^* = [0, 0.500, 0.707, 0.866, 1]$ | $B^* = [0, 0.404, 0.654, 0.904, 1]$ | 96.26 % |
| not $A = 1-[large]$ | $A^* = [1, 0.750, 0.500, 0.250, 0]$ | $B^* = [0.727, 1, 0.500, 0, 0.273]$ | 79.06 % |
| RPCF–FMP–LCM–average | | | 93.15% |
| Fuzzy Rule Antecedent; 1−$B$=[1, 0.750, 0.500, 0.250, 0], Consequent; 1−$A$=[1, 0.75, 0.5, 0.25, 0] | | | |
| The Given Premise; $B^*(y)$ | | Conclusion; $A^*(x)=?$ | RPCF |
| not $B = 1-[large]$ | $B^* = [1, 0.750, 0.500, 0.250, 0]$ | $A^* = [1, 0.750, 0.500, 0.250, 0]$ | 100 % |
| not very $B = 1-[large]^2$ | $B^* = [1, 0.938, 0.750, 0.438, 0]$ | $A^* = [1, 0.913, 0.663, 0.413, 0]$ | 97.28 % |
| not more or less $B$ | $B^* = [1, 0.500, 0.293, 0.134, 0]$ | $A^* = [1, 0.596, 0.346, 0.096, 0]$ | 96.26 % |
| $B = [large]$ | $B^* = [0, 0.250, 0.500, 0.750, 1]$ | $A^* = [0.273, 0, 0.500, 1, 0.727]$ | 79.06 % |
| RPCF–FMT–LCM–average | | | 93.15% |

**Table 14.** Comparison of [13,17]'s and proposed method LCM with respect to the reductive property and fuzzy control

| No | Fuzzy Reasoning Method | | Reductive Property For FMP | Reductive Property For FMT | Reductive Property | Fuzzy Controllability |
|---|---|---|---|---|---|---|
| 1 | Mizumoto [17] | $R_{ss}$ | 100 % | 100 % | 100 % | No |
| 2 | Proposed | LCM | 93.15% | 93.15% | 93.15% | Yes |
| 3 | Mizumoto [17] | $R_{sg}$ | 100 % | 75% | 87.5% | No |
| 4 | Mizumoto [17] | $R_s$ | 75% | 75% | 75% | No |
| 5 | Mizumoto [17] | $R_{gg}$ | 75% | 0 % | 37.5% | No |
| 6 | Mizumoto [17] | $R_{gs}$ | 75% | 0 % | 37.5% | No |
| 7 | Mizumoto [17] | $R_g$ | 50 % | 0 % | 25% | No |

| 8 | Mamdani [13] | $R_c$ | 25% | 0 % | 12.5% | Yes |

Analysis for fuzzy reasoning methods presented in [13,17,32] and proposed method LCM can be summarized as follows.

**on 5.11** As know from Table 11-14 the reductive property of the fuzzy reasoning method $R_{ss}$ is more than

**Proposition 5.11** As know from Table 11-14 the reductive property of the fuzzy reasoning method $R_{ss}$ is more than proposed LCM, but cannot be applied to fuzzy control. And the reductive property of $R_s$, $R_g$, $R_{sg}$, $R_{gg}$, $R_{gs}$, and $R_c$ are less than proposed LCM, respectively. And since the fuzzy reasoning methods $R_m$, $R_a$, $R_\#$, $R_\Delta$, $R_\square$ and $R_*$ do not satisfy the reductive property, which cannot be applied to the practical problems, for example fuzzy control.

**Proposition 5.12** Our proposed method LCM has not only high reductive property but also fuzzy controllability.

# 6. Conclusions

Firstly, in this paper we prroposed a novel original method of fuzzy approximate reasoning that can open a new direction of research in the uncertainty inference of AI and CI, which is based on distance measure, concretely, an extended distance measure by the least common multiple(LCM). our proposed fuzzy approximate reasoning method LCM based on an least common multiple is an inference one for the SISO fuzzy system with discrete fuzzy set vectors of equal or different dimensions between the antecedent and consequent of fuzzy rule. We call it LCM method. LCM method is consisted of two part, i.e., FMP−LCM, and FMT−LCM.

Secondly, We proposed and proved 4 theorems with respect to the reductive property and information loss. And we proposed and proved 6 theorems with respect to fuzzy controllability of the several fuzzy reasoning methods based on the fuzzy relation and our proposed method based on LCM.

Thirdly, we compared the reductive properties for 5 fuzzy reasoning methods, i.e., CRI, TIP, QIP, AARS, and an our proposed LCM with respect to FMP and FMT. The experimental results highlight that the proposed approximate reasoning method LCM is comparatively clear and effective, and in accordance with human thinking than the existing fuzzy reasoning methods.

**References**
[1] BLOC L., BOROWIK P., *Many–Valued Logic*. Springer–Verlag. Warsawa, 1993.
[2] BUCKLEY J., HAYASHI Y., *Can approximate reasoning be consistent?*. Fuzzy Sets and Systems. **65**(1994), 13–18.
[3] CHEN S. M., *A new approach to handling fuzzy decision–making problems*. IEEE Trans. Syst. Man Cybern. (**18**)(1988), 1012–1016.
[4] CHEN S. M., *A weighted fuzzy reasoning algorithm for medical diagnosis*. Decis. Support Syst. **11**(1994), 37–43.
[5] CHEN Q., KAWASE S., *Fuzzy Distance Between Fuzzy Values,* Japan Society for Fuzzy and Systems, **10(3)**(1998), 173−176.
[6] DEER P.J., EKLUND P.W., and NORMAN B.D., *A Mahalanobis Distance Fuzzy Classifier,* Proc. 1996 Australian New Zealand Conf. on Intelligent Information Systems, **18(11)**(1996), 1−4.
[7] DENG G., JIANG Y., *Fuzzy reasoning method by optimizing the similarity of truth–tables*. Information Sciences, **288**(2014), 290–313.
[8] DUBOIS D., PRADE H., *Fuzzy Sets and Systems*. Academic Press, New York, 1980.
[9] HE Y. S., QUAN H. J., DENG H. W., *An Algorithm of General Fuzzy Inference With The Reductive Property*. Computer Science, **34**(2007), 145–148.
[10] HE Y. S., QUAN H.J., *Study on the results of triple I method, Computer Sciences,* **39**(2012), 248–250.
[11] JANTZEN J., *Foundations of Fuzzy Control. John Wiley and Sons*, Chichester, 2007.
[12] KWAK S.I., KIM K.J., KIM U.H., RI G.C., KIM H.C., RYU U.S., *Reductive Property of New Fuzzy Reasoning Method based on Distance Measure*, Artificial Intelligence, arXiv: 1809.05001, Cornell University, US, 2018.09.13 Published.
[13] Chung-Jin KWAK, Kwang-Chol RI, Son-Il KWAK, Kum-Ju KIM, Un-Sok RYU, O-Chol KWON, and Nam-Hyok KIM, *Fuzzy Modus Ponens and Fuzzy Modus Tollens based on Fuzzy Moving Distance in SISO Fuzzy System*, Romanian Journal of Information Science and Technology, Accepted for publication, 29th April, 2020.
[14] Son-Il KWAK, Un-Sok RYU, Kum-Ju KIM, Myong-Hye JO, *A fuzzy reasoning method based on compensating operation and its application to fuzzy systems*, Iranian Journal of Fuzzy Systems, **16(3)**(2019), 17−34.
[15] LIU H. W., *Notes on triple I method of fuzzy reasoning*. Computers and Mathematics with Applications, **58**(2009), 1598–1603.
[16] LUO M. X., and YAO N., *Triple I algorithms based on Schweizer–Sklar operators in fuzzy reasoning*. International Journal of Approximate Reasoning, **54**(2013), 640–652.
[17] LUO M., and ZHOU K., *Logical foundation of the quintuple implication inference methods*, International Journal of Approximate Reasoning, **101**(2018), 1−9
[18] MAMDANI E. H., *Application of fuzzy logic to approximate reasoning using linguistic systems*. IEEE Transactions on Computers, **26**(1977), 1182–1191.
[19] MIZUMOTO M., *Fuzzy inference using max –^ composition in the compositional rule of inference*. In: M.M. Gupta, E. Sanchez (Eds.), Approximate Reasoning in Decision Analysis, North Holland, Amsterdam, (1982), 67–76.
[20] MIZUMOTO M., *Fuzzy conditional inference under max– composition*. Information Sciences, **27**(1982), 183–209.
[21] MIZUMOTO M., *Extended fuzzy reasoning*. In: M.M. Gupta, A. Kandel, W. Bandler, J.B. Kiszka (Eds.), Approximate Reasoning

in Expert Systems, North Holland, Amsterdam, 1985.
[22] MIZUMOTO M., *Improvement of fuzzy control*. 3[th] Fuzzy System Symposium, (1987), 153–158.
[23] MIZUMOTO M., VIMMERMANN J., *Comparison of fuzzy reasoning methods*. Fuzzy Sets and Systems, **8**(1982), 253–283.
[24] RAHA S., PAL N. R., RAY K. S., *Similarity based approximate reasoning: Methodology and Application,* IEEE Transactions on Systems Man Cybernetics, A, **32(4)**(2002), 541-547.
[25] RAHA S., HOSSAIN A., GHOSH S., *Similarity based approximate reasoning: fuzzy control*, Journal of Applied Logic, **6**(2008), 47–71.
[26] TAKACS M., *Modified Distance Based Operators in Fuzzy Approximate Reasoning*, ICM 2006 IEEE 3[rd] International Conference on Mechatronics, (2006), 297–299.
[27] TURKSEN I. B., ZHONG Z., *An approximate analogical reasoning scheme based on similarity measures and interval valued fuzzy sets*. Fuzzy Sets and Systems, **34** (1990), 323–346.
[28] TURKSEN I. B., ZHONG Z., *An approximate analogical reasoning approach based on similarity measures*. IEEE Trans. Syst. Man Cybern. **18**(1988), 1049–1056.
[29] WANG D. G., MENG Y. P., LI H. X., *A fuzzy similarity inference method for fuzzy reasoning*, Computers and Mathematics with Applications, **56**(2008), 2445–2454.
[30] WANG G. J., *On the logic foundations of fuzzy modus ponens and fuzzy modus tollens,* Journal Fuzzy Mathematics, **5**(1997), 229.
[31] WANG G. J., *Logic on certain algebras* (I). J. Shaanxi Normal University, **25(1)** (1997), 11-17.
[32] WANG G. J., *The triple I method with total inference rules of fuzzy reasoning*. Science in China, E **29**(1999), 43–53.
[33] WANG G. J*., Formalized theory of general fuzzy reasoning*. Information Sciences, **160**(2004), 251–266.
[34] YEUNG D. S., TSANG E.C.C., *A comparative study on similarity based fuzzy reasoning methods*. IEEE Trans. on Systems, Man, and Cybernetics Part B: Cybernetics, **27**(1997), 216–227.
[35] YEUNG D. S., TSANG E.C.C., *Improved fuzzy knowledge representation and rule evaluation using fuzzy Petri nets and degree of subsethood*. Intelligence Systems, **9**(1994), 1083–1100.
[36] YUAN X. H., LEE E. S., *Comments on "Notes on triple I method of fuzzy reasoning"*. Computers and Mathematics with Applications, **58**(2009), 1604–1607.
[37] ZHAO Z. H., LI Y. J., *Reverse triple I method of fuzzy reasoning for the implication operator RL*. Computers and Mathematics with Applications, **53**(2007), 1020–1028.
[38] ZADEH L. A., *The concept of a linguistic variable and its applications to approximate reasoning I*, Information Sciences, **8** (1975), 199–249.
[39] ZHENG M. C., SHI Z. K., LIU Y., *Triple I method of approximate reasoning on Atanassovs intuitionistic fuzzy sets.* International Journal of Approximation Reasoning. (2014). http://dx.doi.org/10.1016 /j.ijar.2014.01.001
[40] ZHOU B., XU G, LI S. J., *The Quintuple Implication Principle of fuzzy reasoning*. Information Sciences, **297**(2015), 202–215.

Il-Myong SON: Faculty of Information Science, KIM IL SUNG University, Pyongyang, D P R KOREA
Son-Il KWAK[*]: Faculty of Information Science, KIM IL SUNG University, Pyongyang, D P R KOREA
[*]Corresponding Author, *E–mail address*: si.kwak@ryongnamsan.edu.kp
Myong-Ok CHOE: Faculty of Information Science, KIM IL SUNG University, Pyongyang, D P R KOREA